\documentclass{article}
\usepackage{spconf,amsmath,graphicx}
\usepackage{multirow,amsfonts,cite,array,url}
\usepackage{color}

\title{MM-DFN: Multimodal Dynamic Fusion Network For Emotion Recognition in Conversations}
\name{
    Dou Hu$^1$ \thanks{Corresponding author. Email: HUDOU470@pingan.com.cn} \qquad
    Xiaolong Hou$^1$ \qquad 
    Lingwei Wei$^{2}$ \qquad 
    Lianxin Jiang$^1$ \qquad 
    Yang Mo$^1$}
\address{
    $^1$ Ping An Life Insurance Company of China, Ltd. \\
    $^2$ Institute of Information Engineering, Chinese Academy of Sciences  
}

\begin{document}
%
\maketitle
\begin{abstract}

  Emotion Recognition in Conversations (ERC) has considerable prospects for developing empathetic machines. For multimodal ERC, it is vital to understand context and fuse modality information in conversations. Recent graph-based fusion methods generally aggregate multimodal information by exploring unimodal and cross-modal interactions in a graph. However, they accumulate redundant information at each layer, limiting the context understanding between modalities. In this paper, we propose a novel Multimodal Dynamic Fusion Network (MM-DFN) to recognize emotions by fully understanding multimodal conversational context. Specifically, we design a new graph-based dynamic fusion module to fuse multimodal context features in a conversation.  The module reduces redundancy and enhances complementarity between modalities by capturing the dynamics of contextual information in different semantic spaces. Extensive experiments on two public benchmark datasets demonstrate the effectiveness and superiority of the proposed model. 
\end{abstract}


\section{Introduction}
\label{sec:intro}
Emotion Recognition in Conversations (ERC) aims to detect emotions in each utterance of the conversation.  
It has considerable prospects for developing empathetic machines \cite{DBLP:journals/inffus/MaNXC20}.
This paper studies ERC under a multimodal setting,
\textit{i.e.}, acoustic, visual, and textual modalities. 
A conversation often contains rich contextual clues \cite{DBLP:conf/acl/PoriaHMNCM19,DBLP:conf/acl/HuWH20}, which are essential for identifying emotions.
The key success factors of multimodal ERC are accurate context understanding and multimodal fusion.
Previous context-dependent works \cite{DBLP:conf/aaai/MajumderPHMGC19,DBLP:conf/emnlp/GhosalMPCG19,DBLP:conf/acl/HuWH20} model conversations as sequence or graph structures to explore contextual clues within a single modality. 
Although these methods can be naturally extended multimodal paradigms by performing early/late fusion such as \cite{DBLP:conf/acl/PoriaCHMZM17, DBLP:conf/emnlp/HazarikaPMCZ18, DBLP:conf/icmcs/FuOWGSLD21}, it is difficult to capture contextual interactions between modalities, which limits the utilization of multiple modalities.
Besides, some carefully-designed hybrid fusion methods \cite{DBLP:conf/emnlp/ZadehCPCM17,DBLP:conf/acl/MorencyLZLSL18,DBLP:conf/aaai/ZadehLMPCM18,DBLP:conf/mm/ChenSOLS21} focus on the alignment and interaction between modalities in isolated or sequential utterances. 
These methods  
ignore complex interactions between utterances, 
resulting in leveraging context information in conversations insufficiently.


Recent remarkable works \cite{DBLP:conf/acl/HuLZJ20,DBLP:conf/icassp/LiuCWLFGD21} model unimodal and cross-modal interactions in a graph structure, which provides complementarity between modalities for tracking emotions. 
However, these graph-based fusion methods aggregate contextual information in a specific semantic space at each layer, gradually accumulating redundant information. 
It limits context understanding between modalities. 
The contextual information continuously aggregated can be regarded as specific views where each view can have its individual representation space and dynamics.
We believe that modeling these dynamics of contextual information in different semantic spaces can reduce redundancy and enhance complementarity, accordingly boosting context understanding between modalities.

In this paper, we propose a novel Multimodal Dynamic Fusion Network (MM-DFN) to recognize utterance-level emotion by sufficiently understanding multimodal conversational context.
Firstly, we utilize a modality encoder to 
track speaker states and context in each modality. 
Secondly, 
inspired by \cite{DBLP:journals/neco/HochreiterS97,DBLP:conf/icml/ChenWHDL20},
we improve the graph convolutional layer  \cite{DBLP:conf/iclr/KipfW17} with gating mechanisms and design a new Graph-based Dynamic Fusion (GDF) module to fuse multimodal context information. 
The module utilizes graph convolution operation to aggregation context information of both inter- and intra-modality in a specific semantic space at each layer.
Meanwhile, the gating mechanism is used to learn the intrinsic sequential patterns of contextual information in adjacent semantic space. The GDF module can control information flow between layers, reducing redundancy and promoting the complementarity between modalities.
The stack of GDFs can naturally fuse multimodal context features by embedding them into a dynamic semantic space. 
Finally, 
an emotion classifier is used to predict the emotion label of the utterance.

\begin{figure*}[t]
  \centering
  \includegraphics[width=0.93\linewidth]{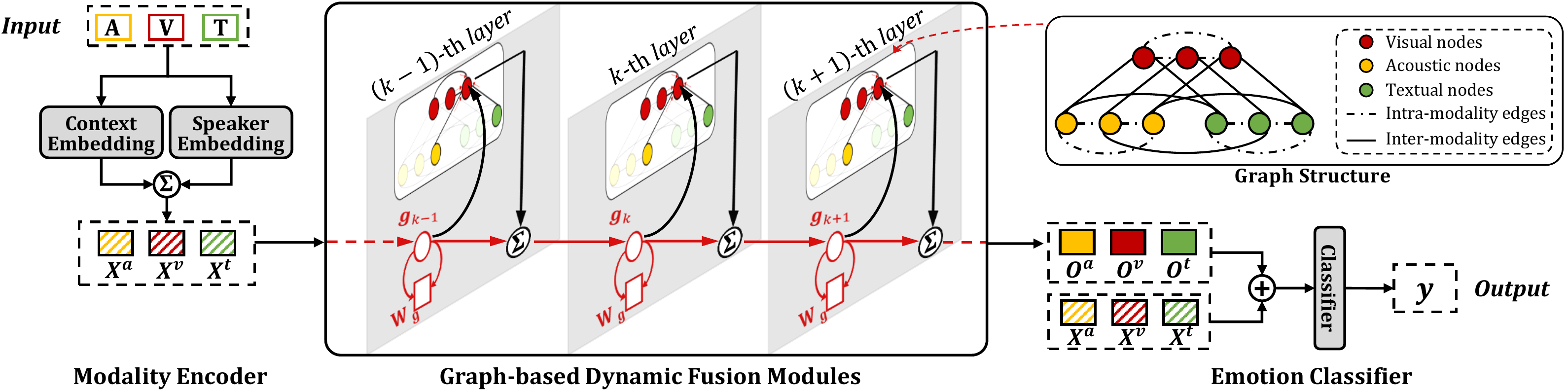}
  \caption{The architecture of the proposed MM-DFN. Given input multimodal features, modality encoder first captures features of context and speaker in each modality.
  Then, in each conversation, we construct the fully connected graph in each modality, and connect nodes corresponding to the same utterance between different modalities.
  Based on the graph, graph-based dynamic fusion modules are stacked to fuse multimodal context features, dynamically and sequentially.  
    Finally, based on the concatenation of features, an emotion classifier is applied to identify emotion label of each utterance.
 }
  \label{fig:overall}
\end{figure*}


We conduct a series of experiments on two public benchmark datasets, {\it i.e.,} {\it IEMOCAP} and {\it MELD}. Results consistently demonstrate that MM-DFN significantly outperforms comparison methods.  
The main contributions are summarized as follows:
1) We propose a novel MM-DFN to facilitate multimodal context understanding for ERC. 
2) We design a new graph-based dynamic fusion module to fuse multimodal conversational context.
This module can reduce redundancy and enhance complementarity between modalities.
3) Extensive experiments on two benchmark datasets demonstrate the effectiveness and superiority of the proposed model\footnote{The code is available at {https://github.com/zerohd4869/MM-DFN}}.

%




\section{Methodology}

Formally, given a conversation $U=[u_1, ..., u_N]$, $u_i = \{ \mathbf{u}^a_i,  \mathbf{u}^v_i, \mathbf{u}^t_i \}$, where $N$ is the number of utterances. 
$\mathbf{u}^a_i, \mathbf{u}^v_i, \mathbf{u}^t_i$ denote the raw feature representation of $u_i$ from the acoustic, visual, and textual modality, respectively.
There are $M$ speakers $ P =\{p_1, ..., p_M\}(M \geq 2)$.
Each utterance $u_i$ is spoken by the speaker $p_{\phi(u_i)}$, where $\phi$ maps the index of the utterance into the corresponding speaker.
Moreover, we define $U_{{\lambda}}$ to represent the set of utterances spoken by the party $p_{\lambda}$.
  $U_{{\lambda}} = \{ u_i | u_i \in U \text{ and }  u_i \text{ spoken by }   p_{\lambda}, \text{ } \forall i \in [1, N]   \}, \lambda \in [1,M]$.
The goal of multimodal ERC is to predict the emotion label $y_i$ for each utterance $u_i$ from pre-defined emotions $\mathcal{Y}$.

In this section, we propose a novel Multimodal Dynamic Fusion Network (MM-DFN) to fully understand the multimodal conversational context for ERC,
as shown in Fig.~\ref{fig:overall}.


\subsection{Modality Encoder}
To capture context features for the textual modality, we apply a bi-directional gated recurrent unit (BiGRU); for the acoustic and visual modalities, we apply a  fully connected network. The context embedding can be computed as: 
\begin{equation}
 \begin{split}
  & \mathbf{c}^{\varsigma}_i = \mathbf{W}^{\varsigma}_c \mathbf{u}_{i}^{\varsigma} + \mathbf{b}^{\varsigma}_c, \varsigma \in \{ a, v\}, \\
  & \mathbf{c}^{t}_i, \mathbf{h}^{c}_{i}  = {\overleftrightarrow{GRU}_c}(\mathbf{u}^t_{i}, \mathbf{h}^{c}_{i-1}), \\
\end{split}
\end{equation} 
where $\overleftrightarrow{GRU}_c$ is a BiGRU to obtain context embeddings and $\mathbf{h}^c_{i}$ is the hidden vector.  
$\mathbf{W}^a_c, \mathbf{W}^v_c, \mathbf{b}^a_c, \mathbf{b}^v_c$ are trainable parameters.
Considering the impact of speakers in a conversation, we also employ a shared-parameter BiGRU to encode different contextual information from multiple speakers:
\begin{equation}
\resizebox{0.88\linewidth}{!}{$
\mathbf{s}^{\delta}_i, \mathbf{h}^{s}_{\lambda, j}  = {\overleftrightarrow{GRU}_s}(\mathbf{u}^{\delta}_{i}, \mathbf{h}^{s}_{\lambda,j-1}), j \in [1,|U_{\lambda}|], \delta \in \{ a, v, t \},
$}
\end{equation}  
where $\overleftrightarrow{GRU}_s$ indicates a BiGRU to obtain speaker embeddings. 
$\mathbf{h}^s_{\lambda, j}$ is the $j$-th hidden state of the party $p_\lambda$. $\lambda = \phi(u_i)$. 
$U_\lambda$ refers to all utterances of $p_\lambda$ in a conversation.  

\subsection{Graph-based Dynamic Fusion Modules}
\subsubsection{Graph Construction}
Following \cite{DBLP:conf/acl/HuLZJ20}, we build an undirected graph to represent a conversation, denoted as $\mathcal{G} = (\mathcal{V}, \mathcal{E})$.
$\mathcal{V}$ refers to a set of nodes. Each utterance can be represented by three nodes for differentiating acoustic, visual, and textual modalities. Given $N$ utterances, there are $3N$ nodes in the graph.
We add both context embedding and speaker embedding to initialize the embedding of nodes in the graph:
\begin{equation}
  \mathbf{x}^{\delta}_i = \mathbf{c}^{\delta}_i + \gamma^{\delta} \mathbf{s}^{\delta}_i, \delta \in \{ a, v, t\},
\end{equation} 
where $\gamma^{a}, \gamma^{v}, \gamma^{t}$ are trade-off hyper-parameters.
$\mathcal{E}$ refers to a set of edges, which are built based on two rules. 
The first rule is that any two nodes of the same modality in the same conversation are connected.
The second rule is that each node is connected with nodes corresponding to the same utterance but from different modalities.
Following \cite{DBLP:conf/textgraphs/SkianisMV18}, edge weights are computed as:
$\mathbf{A}_{ij} = 1 - \frac{\text{arccos}(\text{sim}(\mathbf{x}_i,\mathbf{x}_j))}{\pi},$
where $\text{sim}(\cdot)$ is cosine similarity function.

\subsubsection{Dynamic Fusion Module}
Based on the graph, we improve 
\cite{DBLP:conf/iclr/KipfW17}
with gating mechanisms to fuse multimodal context features in the conversation. 
We utilize graph convolution operation to aggregate context information of both inter- and intra-modality in a specific semantic space at each layer.
Meanwhile, inspired by \cite{DBLP:journals/neco/HochreiterS97}, we leverage gating mechanisms to learn  intrinsic  sequential patterns of contextual  information in different semantic spaces.
The updating process using gating mechanisms is defined  as:
\begin{equation}
  \resizebox{0.8\linewidth}{!}{$
  \begin{split}
    \mathbf{\Gamma}^{(k)}_{\varepsilon} &= \sigma(\mathbf{W}^g_{{\varepsilon}} \cdot [\mathbf{g}^{(k-1)}, {\mathbf{H}'}^{(k-1)}] + \mathbf{b}^g_{\varepsilon} ),  {\varepsilon}=\{ u, f, o \}, \\  
    \tilde{\mathbf{C}}^{(k)} &= \text{tanh}(\mathbf{W}^g_C \cdot [\mathbf{g}^{(k-1)}, {\mathbf{H}'}^{(k-1)}] + \mathbf{b}^g_C ), \\
    \mathbf{C}^{(k)} &=  \mathbf{\Gamma}^{(k)}_f \odot \mathbf{C}^{(k-1)} + \mathbf{\Gamma}^{(k)}_u  \odot  \tilde{\mathbf{C}}^{(k)}, \ \mathbf{g}^{(k)} = \mathbf{\Gamma}^{(k)}_o \odot \text{tanh}( \mathbf{C}^{(k)}  ),
  \end{split}
  $}
\end{equation}
where $\mathbf{\Gamma}^{(k)}_u, \mathbf{\Gamma}^{(k)}_f, \mathbf{\Gamma}^{(k)}_o$ refer to the update gate, the forget gate, and the output gate in the $k$-th layer, respectively. 
$\mathbf{g}^{(0)}$ is initialized with zero. 
$\mathbf{W}^g_{\Gamma}, \mathbf{b}^g_{\Gamma}$ are learnable parameters. 
$\sigma(\cdot)$ is a sigmoid function.
$\tilde{\mathbf{C}}^{(k)}$ stores contextual information of previous layers.
The update gate $\mathbf{\Gamma}^{(k)}_u$ controls what part of the contextual information is written to the memory, while the forget gate $\mathbf{\Gamma}^{(k)}_f$  decides what redundant information in ${\mathbf{C}}^{(k)}$ is deleted.
The output gate $\mathbf{\Gamma}^{(k)}_o$ reads selectively for passing into a graph convolution operation. 
 Following \cite{DBLP:conf/icml/ChenWHDL20}, the modified convolution operation can be defined as: 
\begin{equation}  
\resizebox{0.99\linewidth}{!}{$
\begin{split}
    \mathbf{H}^{(k)} = \text{ReLU} \left(
    (  (1-\alpha) \tilde{\mathbf{P}} {\mathbf{H}'}^{(k-1)} + \alpha \mathbf{H}^{(0)} ) 
    (  (1-\beta_{k-1}) \mathbf{I}_n + \beta_{k-1} \mathbf{W}^{(k-1)} ) 
     \right), \\ 
\end{split}
$}
\end{equation}
where $\tilde{\mathbf{P}} = \tilde{\mathbf{D}}^{-1/2} \tilde{\mathbf{A}} \tilde{\mathbf{D}}^{-1/2}$ is the graph convolution matrix with the renormalization trick. 
$\alpha, \beta_{k}$ are two hyperparameters. $\beta_{k}=log(\frac{\rho}{k} + 1)$. $\rho$ is also a hyperparameter.  $\mathbf{W}^{(k)}$ is the weight matrix.
${\mathbf{H}}^{(0)}$ is initialized with $ \mathbf{X}^a, \mathbf{X}^v, \mathbf{X}^t$.
$\mathbf{I}_n$ is an identity mapping matrix.
Then, the output of $k$-th layer can be computed as,
  ${\mathbf{H}'}^{(k)} = {\mathbf{H}}^{(k)} + \mathbf{g}^{(k)}$.


\subsection{Emotion Classifier} \label{sec:emo}
After the stack of $K$ layers, representations of three modalities for each utterance $i$ can be refined as $\mathbf{o}^a_i, \mathbf{o}^v_i, \mathbf{o}^t_i$. 
Finally, a classifier is used to predict the emotion of each utterance:
\begin{equation}
    \hat{\mathbf{y}}_i = \text{Softmax}(\mathbf{W}_z  [\mathbf{x}^a_i;\mathbf{x}^v_i;\mathbf{x}^t_i; \mathbf{o}^a_i;\mathbf{o}^v_i;\mathbf{o}^t_i] + \mathbf{b}_z),    
\end{equation} 
where $\mathbf{W}_z$ and $\mathbf{b}_z$ are trainable parameters.
We apply cross-entropy loss  along with L2-regularization to train the model:
\begin{equation}
\resizebox{0.8\linewidth}{!}{$
    \mathcal{L} = - \frac{1}{\sum_{l=1}^L \tau(l)} \sum_{i=1}^{L} \sum^{\tau(i)}_{j=1} {\mathbf{y}}^l_{i,j} log (\hat{\mathbf{y}}^l_{i,j}) + \eta \| \Theta \|_2, 
$}
\end{equation}    
where $L$ is the total number of samples in the training set. $\tau(i)$ is the number of utterances in sample $i$. $\mathbf{y}^l_{i,j}$ and $\hat{\mathbf{y}}^l_{i,j}$ denote the one-hot vector and probability vector for emotion class $j$ of utterance $i$ of sample $l$, respectively.
$\Theta$ refers to  all trainable parameters. $\eta$ is the L2-regularization weight.


\section{Experiments}

\begin{table*}[t]
  \centering
      \resizebox{1.0\linewidth}{!}{$
  \begin{tabular}{l|cccccc|cc||ccccc|cc}
  \hline
  \multicolumn{1}{c|}{\multirow{2}{*}{\textbf{Methods}}} & \multicolumn{8}{c||}{\textbf{IEMOCAP}} & \multicolumn{7}{c}{\textbf{MELD}} 
  \\ 
  \cline{2-16}
   & \textit{Happy} &\textit{Sad} & \textit{Neutral} & \textit{Angry} & \textit{Excited} & \textit{Frustrated} & {{Acc}} & {{w-F1}} 
   & \textit{Neutral} & \textit{Surprise} & \textit{Sadness}  
   &  \textit{Happy}
   &\textit{Anger}  & {Acc} & {{w-F1}}    \\
  \hline
    TFN \cite{DBLP:conf/emnlp/ZadehCPCM17}
    & 37.26 & 65.21 & 51.03 & 54.64 & 58.75 & 56.98 & 55.02 & 55.13
    & 77.43 & 47.89 & 18.06 & 51.28 & 44.15 & 60.77 & 57.74 \\ 
    LMF \cite{DBLP:conf/acl/MorencyLZLSL18}
    & 37.76 & 66.53 & 52.39 & 57.53 & 58.41 & 59.27 & 56.50 & 56.49
    & 76.97 & 47.06 & 21.15 & 54.20 & 46.64 & 61.15 & 58.30 \\
    MFN \cite{DBLP:conf/aaai/ZadehLMPCM18}    
    &  48.19	&   73.41	&   56.28 &     63.04 &     64.11 &	    61.82 &	    61.24	&   61.60	
    &  77.27 & 48.29 &  23.24 & 52.63 & 41.32 &   60.80	&   57.80   \\
    bc-LSTM \cite{DBLP:conf/acl/PoriaCHMZM17} 
    &  33.82   & 	78.76   & 	56.75 & 	64.35 & 	60.25 & 	60.75 & 	60.51   & 	60.42   
    &  75.66 & 48.57 & 22.06 & 52.10  &  44.39 &  59.62   & 	57.29   \\ 
    ICON \cite{DBLP:conf/emnlp/HazarikaPMCZ18}  
    & 32.80    & 	74.40   & 	60.60 & 	68.20 & 	68.40 & 	66.20 & 	64.00   & 	63.50   
    & -  & -  & -  & - & -  & - & -    \\ 
    DialogueRNN \cite{DBLP:conf/aaai/MajumderPHMGC19} 
    & 32.20    & 	80.26   & 	57.89 & 	62.82 & 	73.87 & 	59.76 & 	63.52   & 	62.89   
    & 76.97 & 47.69 & 20.41 & 50.92 & 45.52 & 60.31 & 57.66 \\
    DialogueCRN \cite{DBLP:conf/acl/HuWH20} 
    & \textbf{53.23} & \textbf{83.37} & 62.96 & 66.09 & 75.40 & 66.07 & 67.16 & 67.21
    & 77.01 & 50.10 & 26.63 & 52.77 & 45.15 & 61.11 & 58.67  \\
    DialogueGCN \cite{DBLP:conf/emnlp/GhosalMPCG19} 
    & 51.57	&   80.48	&   57.69 & 	53.95 &	    72.81 &	    57.33 &	    63.22	&   62.89	
    & 75.97 & 46.05 & 19.60 & 51.20 & 40.83 & 	58.62   & 	56.36   \\
    MMGCN \cite{DBLP:conf/acl/HuLZJ20}
    & 45.14    & 	77.16   & 	64.36 & 	68.82 & 	74.71 & 	61.40 & 	66.36   & 	66.26   
    & 76.33 & 48.15 & \textbf{26.74} & 53.02 & 46.09 & 60.42 & 58.31 \\
    \hline 
    \textbf{MM-DFN}   
    & 42.22    & 	78.98   & 	\textbf{66.42}$^{*}$ & 	\textbf{69.77}$^{*}$ & 	\textbf{75.56}$^{*}$ & 	\textbf{66.33}$^{*}$ & 	\textbf{68.21}$^{*}$   & 	\textbf{68.18}$^{*}$   
    & \textbf{77.76}$^{*}$   &   \textbf{50.69}$^{*}$  &  {22.93}  &  \textbf{54.78}$^{*}$  &  \textbf{47.82}$^{*}$ & 	\textbf{62.49}$^{*}$   & 	\textbf{59.46}$^{*}$   \\
  \hline
  \end{tabular}    
  $}
  \caption{
  Results under the multimodal setting (A+V+T).
  We present the overall performance of Acc and w-F1, which mean the overall accuracy score and weighted-average F1 score, respectively.
  We also report F1 score per class, except two classes (i.e. \textit{Fear} and \textit{Disgust}) on MELD, whose results are not statistically significant due to the smaller number of training samples.
  Best results are highlighted in bold.  $^{*}$ represents statistical significance over state-of-the-art scores under the paired-$t$ test ($p < 0.05$).
%
} \label{tab:result}
\end{table*}

\begin{table}[t]
  \centering
      \resizebox{1.0\linewidth}{!}{$
    \begin{tabular}{p{6cm}|p{1.9cm}<{\centering}|p{1.9cm}<{\centering}}
      \hline
      \multicolumn{1}{c|}{\multirow{1}{*}{\textbf{Methods}}} 
      & \multicolumn{1}{c|}{\multirow{1}{*}{\textbf{IEMOCAP}}}  
      & \multicolumn{1}{c}{\multirow{1}{*}{\textbf{MELD}} } \\ 
      \hline
      \text{MM-DFN}                                     & 	{68.18}       & 	{59.46}                 \\ 
      \  { } - w/o GDF \ - w Speaker  \ \ \ \ - w Context          & 	63.80         &	58.50           \\
      \  { } - w GDF \ \ \ \  - w/o Speaker \ - w Context          & 	66.89         & 	58.45       \\
      \  { } - w/o GDF \ - w/o Speaker  \ - w Context        & 	62.90         &	58.50                   \\
      \  { } - w/o GDF \ - w/o Speaker \ - w/o Context      & 	54.81         &	58.08               \\      
      \hline
    \end{tabular}      
  $}
  \caption{
  Ablation results of MM-DFN. We report w-F1 score for both datasets.
  }
\label{tab:abla}
\end{table}

\begin{table}[t]
  \centering
      \resizebox{1.0\linewidth}{!}{$
  \begin{tabular}{p{6cm}|p{1.9cm}<{\centering}|p{1.9cm}<{\centering}} 
  \hline
  \multicolumn{1}{c|}{\multirow{1}{*}{\textbf{Fusion Modules}}} & \multicolumn{1}{c|}{\textbf{IEMOCAP}}  & \multicolumn{1}{c}{\textbf{MELD}} 
  \\ 
  \hline
  {Concat / Gate Fusion}        & 	63.80 / 64.30   & 	58.50 / 57.87   \\ 
  {Tensor / Memory Fusion}	    & 	61.05 / 65.51   & 	58.54 / 58.48   \\ 
{Early / Late Fusion + GCN}    & 64.19 / 65.34  & 58.69 / 58.43 \\
  \hline
  {{Graph-based Fusion (GF)}}        & 67.02     & 58.54     \\ 
    {{ }\  \  - w/o {{Inter-Modal} \   - w {Intra-Modal}}} 	&	66.91 &    58.53 \\
    {{ }\  \  - w {{Inter-Modal}}  \ \ \ \ - w/o  {{Intra-Modal}}}  &	66.11 	&	58.29  \\  
  \hline
  {\textbf{Graph-based Dynamic Fusion (GDF)}}  & 68.18 &  	59.46 \\  
    {{ } \ \  - w/o {{Inter-Modal} \  - w {Intra-Modal}}}  	&	67.82 &	59.15 \\ 
    {{ } \ \  - w {{Inter-Modal}}  \ \ \ \  - w/o  {{Intra-Modal}}}  	&	66.22 	&	58.31 \\  
\hline
  \end{tabular}    
  $}
  \caption{
  Results against different fusion modules. We report w-F1 score for both datasets. 
  } \label{tab:fusion}
\end{table}

\begin{table}[t]
\centering
\resizebox{1.0\linewidth}{!}{$
\begin{tabular}{p{1.6cm}|cc|cc} 
\hline
\multicolumn{1}{c|}{\multirow{2}{*}{\textbf{Modality}}} & \multicolumn{2}{c|}{\multirow{1}{*}{\textbf{IEMOCAP}}}  & \multicolumn{2}{c}{\multirow{1}{*}{\textbf{MELD}} }
\\ 
\cline{2-5}
  & \multirow{1}{*}{{GF}} & \multirow{1}{*}{\textbf{GDF}} & \multirow{1}{*}{{GF}} & \multirow{1}{*}{\textbf{GDF}}  \\ \hline
  A / V / T	 & -   & 	47.79 / 27.46 / 61.07  &  - &	42.72 / 32.34 / 56.95  \\   
  \hline
  A + V	    & 54.73         & 	56.35 & 	42.74 & 	44.67 \\  
  A + T       & 65.03  & 	65.41   & 	57.85 & 	58.34 \\  
  V + T	    & 62.07         & 	62.63 & 	57.78 & 	58.49 \\
  {A + V + T}   & {67.02}       & {68.18} & {58.54}   &   {59.46} \\ 
\hline
\end{tabular} 
$}
\caption{Results of graph-based fusion methods under different modality settings. 
Fusion modules are not used under unimodal types.
We report w-F1 score for both datasets.
} \label{tab:modality}
\end{table}

\subsection{Datasets}
\textit{\textbf{IEMOCAP}} \cite{DBLP:journals/lre/BussoBLKMKCLN08} 
contains dyadic conversation videos between pairs of ten unique speakers. 
It includes 7,433 utterances and 151 dialogues. Each utterance is annotated with one of six emotion labels.
We follow the previous studies \cite{DBLP:conf/emnlp/GhosalMPCG19,DBLP:conf/acl/HuLZJ20} that use the first four sessions for training, use the last session for testing, and randomly extract 10\% of the training dialogues as validation split.
\textit{\textbf{MELD}}
\cite{DBLP:conf/acl/PoriaHMNCM19} 
contains multi-party conversation videos collected from Friends TV series, where two or more speakers are involved in a conversation. 
It contains 1,433 conversations, 13,708 utterances and 304 different speakers. Each utterance is annotated with one of seven emotion labels.
For a fair comparison, we conduct experiments using the pre-defined train/validation/test splits in MELD.
\subsection{Comparison Methods}

\textbf{TFN} \cite{DBLP:conf/emnlp/ZadehCPCM17} and \textbf{LMF} \cite{DBLP:conf/acl/MorencyLZLSL18} 
make non-temporal multimodal fusion by tensor product.
\textbf{MFN} \cite{DBLP:conf/aaai/ZadehLMPCM18} synchronizes multimodal sequences using a multi-view gated memory.
\textbf{bc-LSTM} \cite{DBLP:conf/acl/PoriaCHMZM17} leverages an utterance-level LSTM to capture multimodal features. 
\textbf{ICON} \cite{DBLP:conf/emnlp/HazarikaPMCZ18}, an extension of CMN \cite{DBLP:conf/naacl/HazarikaPZCMZ18}, provides conversational features from modalities by multi-hop memories.
\textbf{DialogueRNN} \cite{DBLP:conf/aaai/MajumderPHMGC19}  
introduces a recurrent network to track speaker states and context during the conversation.
\textbf{DialogueCRN} \cite{DBLP:conf/acl/HuWH20} designs multi-turn reasoning modules to understand conversational context. 
\textbf{DialogueGCN} \cite{DBLP:conf/emnlp/GhosalMPCG19} utilizes graph structures to combine contextual dependencies.
\textbf{MMGCN} \cite{DBLP:conf/acl/HuLZJ20} uses a graph-based fusion module to capture intra- and inter- modality contextual features.
All baselines are reproduced under the same environment, except \cite{DBLP:conf/emnlp/HazarikaPMCZ18}, which is only applicable for dyadic conversation and the results are from the original paper. 
Because \cite{DBLP:conf/aaai/MajumderPHMGC19,DBLP:conf/emnlp/GhosalMPCG19,DBLP:conf/acl/HuWH20} are designed for unimodal ERC, 
a early concatenation fusion is introduced to capture multimodal features in their implementations. %




\textit{\textbf{Implementation Details.}}
Following \cite{DBLP:conf/acl/HuLZJ20}, raw utterance-level features of acoustic, visual, and textual modality are extracted by TextCNN \cite{DBLP:conf/emnlp/Kim14}, \textit{OpenSmile}  \cite{DBLP:journals/speech/SchullerBSS11}, and DenseNet \cite{DBLP:conf/cvpr/HuangLMW17}, respectively. 
We use focal loss \cite{lin2017focal} for training due to the class imbalance. 
The number of layers $K$ are 16 and 32 for IEMOCAP and MELD. 
$\alpha$ is set to 0.2 and $\rho$ is set to 0.5.

\subsection{Experimental Results and Analysis}
\textit{\textbf{Overall Results and Ablation Study.}}
The overall results are reported in Table~\ref{tab:result}.
MM-DFN consistently obtains the best performance over the comparison methods on both datasets, which shows the superiority of our model.
%
Table~\ref{tab:abla} shows ablation studies by removing key components of the proposed model. 
When removing either the graph-based dynamic fusion (GDF) module or speaker embedding (Speaker), the results decline significantly on both datasets. 
When further removing the context embedding (Context), the results decrease further. 
It shows the effectiveness of the three components.

 \textit{\textbf{Comparison with Different Fusion Modules.}}
After the modality encoder, 
we replace GDF with the following six fusion modules:
\textbf{Concat/Gate Fusion}, 
\textbf{Tensor/Memory Fusion}\cite{DBLP:conf/acl/MorencyLZLSL18,DBLP:conf/aaai/ZadehLMPCM18},
\textbf{Early/Late Fusion + GCN }\cite{DBLP:conf/acl/HuLZJ20}, and  
\textbf{Graph-based Fusion} (GF)   \cite{DBLP:conf/acl/HuLZJ20}.
From Table~\ref{tab:fusion}, GF and GDF outperform all fusion modules in the first block since the two graph-based fusion modules sufficiently capture intra- and inter-modality interactions in conversations, which provides complementarity between modalities. 
GDF achieves better performance, reducing  redundancy  and  promoting the  complementarity  between  modalities, which shows the superiority of multimodal fusion. 
Besides, for GF and GDF, we analyze the impact of inter- and intra-modality edges in the graph for fusion.
{Intra-}/{Inter-Modal} refers to building edges according to the first/second rule.
Ignoring any rules can hurt performance in GF and GDF, which shows that modeling contextual interactions of both inter- and intra-modality, can better utilize the complementarity  between modalities.
Compared with GF, GDF obtains a better performance in all variants. 
It shows that GDF can reduce both inter- and intra-modality redundancies and fuse multimodal context better.



 \textit{\textbf{Comparison under Different Modality Settings.}}
Table~\ref{tab:modality} shows the results of MM-DFN and the GF-based variant under different modality settings.
As expected, bimodal and trimodal models outperform the corresponding unimodal models on both datasets. %
Under unimodal types, textual modality performs better than acoustic and visual. 
Under bimodal types, GDF outperforms GF consistently.
It again confirms the superiority of GDF. 
Meanwhile, under acoustic and textual modalities (A+T), both GF and GDF achieve the best performance over other bimodal types, which indicates a stronger complementarity between rich textual semantics and affective audio features. 
GDF can reduce redundancy as well as enhance complementarity between modalities and thus obtain better results. 
Moreover, under acoustic and visual modalities (A+V), GDF outperforms GF by a large margin.
This phenomenon reflects that the acoustic and visual features have high entanglement and redundancy, limiting the performance of GF.
Our GDF encourages disentangling and reduces redundancy by controlling information flow between modalities, accordingly obtaining better fusion representations.

\section{Conclusion}

This paper proposes a Multimodal Dynamic Fusion Network (MM-DFN) to fully understand conversational context for multimodal ERC task. 
A graph-based dynamic fusion (GDF) module is designed to fuse multimodal features in a conversation. The stack of GDFs learns dynamics of contextual information in different semantic spaces, successfully reducing redundancy and enhancing complementarity between modalities. 
Extensive experiments on two benchmark datasets demonstrate the effectiveness and superiority of MM-DFN. 

\end{document}